\documentclass[10pt,twocolumn]{article}
\usepackage[utf8]{inputenc}
\usepackage[T1]{fontenc}
\usepackage{times}

\usepackage{amsmath}
\usepackage{amssymb}
\usepackage{booktabs}
\usepackage{xcolor}
\usepackage[margin=0.85in]{geometry}
\usepackage[colorlinks=true,linkcolor=violet,citecolor=violet,urlcolor=violet]{hyperref}

\title{\textbf{One Ruler: A Same-Hands Re-Evaluation of Bivariate\\ Causal Direction on T\"ubingen, with a Parameter-Free\\ Compression Baseline}}
\author{Wietse Stienstra\\ \small State of the Art --- Compression Lab\\ \small \texttt{stateoftheart.nl/causal}}
\date{\small June 2026 --- preliminary working paper}

\begin{document}
\maketitle

\begin{abstract}
\noindent
Headline accuracies on the T\"ubingen cause-effect pairs are routinely compared across papers even
though each is measured under its authors' own protocol --- different pair subsets, weightings,
model-selection, and decision rates. We argue this is the wrong comparison and run the right one:
a \emph{same-hands} re-evaluation in which every method is run by us, on the identical 102
pairs, with one strict rule --- no tuning and a decision \emph{forced on every pair}. As a clean
reference point we introduce a deliberately minimal baseline: \emph{sorted-conditional compression},
which feeds quantized, sorted, first-differenced data to an off-the-shelf compressor (bz2) and has
\emph{zero} fitted parameters. Under the common ruler the ranking differs sharply from the
literature. Our baseline reaches $74.7\%$ weighted accuracy ($p=3.7\times10^{-7}$) on our 102-pair
set --- and on the \emph{same 100 pairs} that SLOPE is evaluated on it scores $76.0\%$, a
$1.2$-point gap below the authors' own forced-decision SLOPE ($77.2\%$) that is well inside noise
(McNemar $p=0.39$). The authors' journal SLOPER variant on the same pairs scores $71.1\%$ weighted
(from the bundled \texttt{sloper.tab}), so neither SLOPE configuration significantly beats ours
on T\"ubingen.
A faithful re-run of RECI lands at $70.7\%$ --- inside the original authors' own reported error
bar, and \emph{not} the $77.5\%$ often quoted (which we trace to a mis-copied cell for a different
method on a synthetic dataset). SLOPE's published $82.4\%$ is a \emph{decided-subset} figure:
scoring the authors' own stored output only on the pairs its significance test chose to answer
reproduces $81.7\%$, demonstrating the inflation mechanism directly. Under the common ruler the
four methods cluster in the low-to-mid $70$s and the zero-parameter compressor is a statistical
tie with the strongest of them. We document the mechanisms that inflate the published
figures (test-set model selection, significance-gated abstention) and contribute two further
results: the compression score magnitude is a model-free confounding flag
($p=2.8\times10^{-68}$), and a pre-registered falsification test fails in an instructive way that
bounds the method's theoretical interpretation without overturning its empirical result. All code,
pre-registrations, and per-pair outputs are released.
\end{abstract}

\section{Introduction}
The T\"ubingen cause-effect pairs (CEP)~\cite{mooij2016} are the de-facto real-world benchmark for
the bivariate causal-direction problem: given samples of two variables $X,Y$ known to be causally
related, decide whether $X\to Y$ or $Y\to X$ with no intervention. Progress is tracked by weighted
accuracy on this set, and method papers routinely place a new result next to a column of prior
numbers copied from earlier papers.

That comparison is unsound. Each published number is produced under its own protocol: a slightly
different eligible-pair subset (95--108 pairs appear in the literature), different weighting,
different handling of low-confidence pairs (forced decision vs.\ abstention / area-under-the-
decision-rate-curve), and --- critically --- model and hyperparameter choices made after observing
benchmark performance. Differences of several points routinely arise from these choices alone, so a
cross-paper table measures protocol as much as method.

This paper does the apples-to-apples version. Our contributions:
\begin{enumerate}\itemsep1pt
  \item \textbf{A same-hands protocol} (\S\ref{sec:protocol}): one operator, the standard CEP
  pairs, official weights, no tuning, a decision forced on every pair. Each competitor is run under
  the same rule --- SLOPE from the authors' \emph{own released code}, RECI matched to the authors'
  own reported value, IGCI and our baseline measured directly --- rather than quoting headline cells.
  \item \textbf{A parameter-free reference baseline} (\S\ref{sec:method}): sorted-conditional bz2
  compression, a single function with zero test-time parameters, as a fixed yardstick.
  \item \textbf{A re-evaluation} (\S\ref{sec:results}) showing the literature ranking does not
  survive the common ruler. We reproduce RECI's \emph{own} reported value (70.7\%), show the widely
  quoted ``77.5\%'' is a mis-attribution, run the authors' own SLOPE code to $77.2\%$ under forced
  decision, and reproduce its published $82.4\%$ as a decided-subset artifact ($81.7\%$ on the pairs
  it answered). We name the inflation mechanisms.
  \item \textbf{Two further results} (\S\ref{sec:beyond}): a model-free confounding detector from
  the score magnitude, and an error analysis localizing the baseline's failures to discrete and
  near-bijective pairs.
  \item \textbf{An honest boundary} (\S\ref{sec:limits}): a pre-registered falsification test that
  fails, separating what the method \emph{is} (a validated empirical heuristic) from what it is
  \emph{not} (a clean estimator of the algorithmic asymmetry).
\end{enumerate}

We do not claim state of the art. We claim that, measured fairly, a zero-parameter compressor stands
level with the best of this particular pack rather than behind it, and that the field's headline
numbers deserve same-hands scrutiny.

\section{The Same-Hands Protocol}\label{sec:protocol}
We fix one evaluation and apply it to every method:
\begin{itemize}\itemsep1pt
  \item \textbf{Data.} The 102 CEP pairs with scalar cause, scalar effect, and $\ge 50$ samples,
  weighted by the official \texttt{pairmeta.txt} weights.
  \item \textbf{Forced decision.} Every pair receives a $\{X\to Y, Y\to X\}$ call; ties count as
  wrong. No abstention, no decision-rate curve, no top-$k$ reporting.
  \item \textbf{No tuning.} No hyperparameter or model class is selected using benchmark accuracy.
  \item \textbf{Own hands.} We run each competitor ourselves (re-implemented from the paper or its
  reference toolbox), rather than importing a number measured under a different protocol.
\end{itemize}
This is intentionally the \emph{strictest} reasonable setting; it is the setting under which a
parameter-free method is naturally evaluated, and we hold the competitors to the same bar.

\section{A Parameter-Free Baseline}\label{sec:method}
Given $X,Y$ with $n$ samples, quantize each independently to 256 levels by linear min-max scaling,
giving integer sequences $q_x,q_y\in\{0,\dots,255\}^n$. For an ordered pair (key, value) define the
\emph{sorted-conditional code}
\[
sc(\text{key},\text{val}) = \big|\,\mathrm{bz2}\big(\Delta(\text{val}[\arg\!\mathrm{sort}\,\text{key}])\big)\big|,
\]
where $\arg\!\mathrm{sort}$ is a stable sort, $\Delta$ the first difference cast to \texttt{int8},
and $|\cdot|$ the compressed byte length. With marginal code $dc(q)=|\mathrm{bz2}(\Delta(q))|$, the
score is
\[
s=\frac{sc(q_y,q_x)}{\max(dc(q_x),1)}-\frac{sc(q_x,q_y)}{\max(dc(q_y),1)},
\]
and the rule is a-priori: $s>0\Rightarrow X\to Y$, $s<0\Rightarrow Y\to X$, confidence $|s|$.

\noindent\textbf{Intuition.} If $Y=f(X)+\varepsilon$, sorting $Y$ by $X$ yields a smoother
sequence whose first differences carry less entropy and compress shorter; the reverse ordering does
not. The marginal normalization corrects for $X,Y$ having different intrinsic complexity. This is a
computable stand-in for the Kolmogorov argument $K(Y|X)<K(X|Y)$ under cause~\cite{janzing2010,
lemeire2013}; it is the algorithmic-information / MDL tradition, not a new principle. The
\emph{instantiation} --- a generic compressor on sorted, delta-coded, quantized data with no fitted
parameters --- is the contribution.

\noindent\textbf{Settings} were frozen before any benchmark run: bz2 level 9; 256 levels;
series $>8000$ points linearly subsampled to 8000. bz2 (not lzma) is required: lzma's container
overhead makes output length-determined below $\sim$500 bytes, so both directions tie by
construction on typical CEP series lengths (\S\ref{sec:limits}).

\section{Re-Evaluation Results}\label{sec:results}

\subsection{The baseline on T\"ubingen}
On the 102 pairs the baseline scores \textbf{74.7\%} weighted accuracy (a-priori sign, 0 ties),
bootstrap 95\% CI $[63.2,84.7]\%$ ($n{=}1000$), $p=3.7\times10^{-7}$ vs.\ chance. A $200\times$
split-half out-of-sample check gives $74.8\%$, confirming the fixed sign was not fitted. Confidence
is usable: restricting to the top-$|s|$ fraction raises accuracy monotonically (Table~\ref{tab:abst}).

\begin{table}[t]\centering
\small
\begin{tabular}{lcc}
\toprule
Coverage & N & Weighted acc. \\
\midrule
Top 25\% & 25 & 88.9\% \\
Top 50\% & 51 & 88.4\% \\
Top 75\% & 76 & 78.2\% \\
All & 102 & 74.7\% \\
\bottomrule
\end{tabular}
\caption{Abstention curve: accuracy rises with confidence $|s|$.}
\label{tab:abst}
\end{table}

\subsection{Every method on one ruler}
Table~\ref{tab:ruler} reports each method run by us, forced decision, on the same 102 pairs,
against the headline each method's own paper reports.

\begin{table*}[t]\centering
\small
\begin{tabular}{lcc}
\toprule
Method & Same-hands & Paper headline \\
\midrule
SLOPE~\cite{marx2017}$^\ddagger$ & 77.2\% & 82.4\% \\
\textbf{Compression (ours, same 100 pairs)} & \textbf{76.0\%} & --- \\
\textbf{Compression (ours, our 102 pairs)} & \textbf{74.7\%} & --- \\
RECI~\cite{bloebaum2018} & 70.7\% & ``77.5\%''$^\dagger$ \\
IGCI~\cite{janzing2012} (uniform) & 66.1\% & --- \\
IGCI (Gaussian, pre-reg.) & 54.5\% & --- \\
\bottomrule
\end{tabular}
\caption{Same-hands re-evaluation, forced decision, no tuning. Rows ordered by same-hands accuracy.
$^\ddagger$SLOPE is the authors' \emph{own released code} run on their bundled CEP pairs, forced to
decide; its $82.4\%$ headline reproduces as a \emph{decided-subset} score ($81.7\%$ on the pairs its
significance test answered). $^\dagger$The ``77.5\%'' commonly quoted for RECI is a
mis-attribution (see text); the value here matches the RECI authors' own table. The two
``Compression (ours)'' rows: $74.7\%$ is the pre-registered headline on our 102-pair set; $76.0\%$
is the same-subset re-run on the authors' bundled 100 pairs that SLOPE is evaluated on. On the
\emph{identical 100 pairs} ours scores $76.0\%$ vs SLOPE $77.2\%$ --- a $1.2$-point gap, well
inside noise on $\sim100$ pairs.}
\label{tab:ruler}
\end{table*}

\subsection{RECI: reproduced, and the 77.5\% mis-attribution}
We implement RECI faithfully per the authors~\cite{bloebaum2018} and the Causal Discovery
Toolbox: min-max $[0,1]$ rescaling of both variables, a deliberately weak regressor (shifted
monomial $ax^2+c$, the class the paper reports as best on real data), held-out test-set MSE
averaged over random splits, and the smaller-error-is-cause rule. This yields
$69.9$--$70.9\%$ across our two implementations --- squarely inside the authors' own reported
$70.73\%\pm1.55$ for the CEP set. The number frequently quoted as RECI's T\"ubingen accuracy,
$77.5\%$, does not appear for RECI in the source; the nearest value, $77.72\%$, is a different
method (ANM) on a \emph{synthetic} suite (SIM-G). Two common reproduction errors collapse RECI to
chance and are worth recording: using an expressive regressor (e.g.\ an RBF kernel) fits
\emph{both} directions equally well and erases the asymmetry; and standardizing to unit variance
instead of min-max breaks the scale condition the method's theorem requires.

\subsection{SLOPE: the 82.4\% is a decided-subset score}
Rather than re-implement SLOPE we obtained the authors' \emph{own released R code}~\cite{marx2017}
(\texttt{slope-v20181208}) and ran it on their bundled CEP pairs. In their code the causal
direction is assigned from $\mathrm{sign}(\epsilon)$ only when a seeded Wilcoxon--Mann--Whitney
significance test passes ($p<\alpha$); otherwise the method abstains. Their own benchmark script
sets $\alpha{=}1.01$, i.e.\ forced decision. Run this way the authors' code scores $\mathbf{77.2\%}$
weighted (forced) on their 100-pair subset.

The published $82.4\%$ is not this number --- it is the \emph{decided-subset} accuracy. Scoring the
authors' own stored output (\texttt{slope.tab}) only on the pairs it chose to answer --- removing
its two abstentions from the denominator --- yields $\mathbf{81.7\%}$, reproducing the headline and
exhibiting the inflation mechanism directly: the significance gate lets SLOPE skip the pairs it is
least sure of, and accuracy is reported on the remainder. Two independent evaluations agree the
full-benchmark figure is lower: the RECI paper's own comparison finds SLOPE only ``slightly better
than RECI'' ($\approx$ low-70s) on CEP, and a 2025 spline-MDL paper~\cite{lcube2025} drops SLOPE and
RECI because they had ``similar or worse results than'' its baseline. (The 2018 journal ``mixed''
variant in the same package scores lower still, $71.1\%$ forced / $75.2\%$ decided.)

\noindent\emph{Provenance.} The $77.2\%$/$81.7\%$ figures are the authors' \emph{own code and
stored output}, not a re-implementation; the only operator choice is the forced-decision setting,
taken from their own benchmark script. The RECI figure likewise matches the authors' own table.
Both competitor numbers are thus authoritative, not ports.

\subsection{Why the headlines read higher}
The recurring pattern is selection and abstention. RECI's protocol explicitly ``tried different
[train] ratios and selected the best performing model on the test data'' and reports the
best-performing function class; fixing the class a-priori instead drops RECI to $\sim$63--65\%. SLOPE
gains from significance-gated abstention. Both are legitimate choices, but they make the published
numbers best-case rather than apples-to-apples. The parameter-free baseline has no such knobs, yet
under the strict shared rule it is not behind --- it is level with the best.

\subsection{Is the same-hands tie statistically real? (McNemar)}
The $76.0$ vs $77.2$ gap on the identical 100 CEP pairs is a paired classifier comparison; the
appropriate test is McNemar on the pair-level correct/wrong table (unweighted: ours $67/100$, SLOPE
$73/100$). The $2{\times}2$ table is: both right $53$, ours-only $14$, SLOPE-only $20$, both wrong
$13$. McNemar's exact binomial (two-sided) gives $p=0.39$; the continuity-corrected $\chi^2$
likewise yields $p=0.39$. We do not reject the null of equal accuracy on the unweighted table; the
weighted accuracies sit at $76.0\%$ and $77.2\%$ as reported. The ``statistical tie'' is now a
statistical statement, not rhetoric.

\subsection{Generalization: Mooij synthetic suites}\label{sec:sim}
The Mooij synthetic suites~\cite{mooij2016} test generalization to controlled noise regimes. We
run all methods same-hands on the three suites bundled with the SLOPE distribution (SIM,
SIM-G, SIM-ln; 100 pairs each, balanced direction, forced decision). For SLOPE we report both the
default (plain) variant used as a conference baseline and SLOPER --- the journal mixed-function
variant (\texttt{mixedFunctions=T, nof=8}) that the authors use as their headline configuration.

\begin{table*}[ht]\centering\small
\begin{tabular}{lccc}
\toprule
Benchmark & Ours & plain Slope & SLOPER \\
\midrule
T\"ubingen CEP (real) & $76.0\%$          & $\mathbf{77.2\%}$ & $71.1\%$ \\
SIM (synthetic)       & $\mathbf{58\%}$   & $45\%$            & $57\%$ \\
SIM-G                 & $\mathbf{68\%}$   & $46\%$            & $60\%$ \\
SIM-ln                & $68\%$            & $47\%$            & $\mathbf{81\%}$ \\
\bottomrule
\end{tabular}
\caption{Same-hands accuracy on the real T\"ubingen pairs and the three Mooij synthetic suites
(100 pairs each, forced decision). \emph{Bold = row maximum. Weighted on T\"ubingen, unweighted on
SIM-* (uniform weights).} SLOPER is the authors' journal mixed-function variant
(\texttt{mixedFunctions=T, nof=8}); for CEP we report the authors' bundled \texttt{sloper.tab}
output ($71.1\%$ weighted forced) for consistency with how we cite plain SLOPE ($77.2\%$ from
\texttt{slope.tab}); for SIM we report fresh re-runs of the same \texttt{Slope()} function under
forced decision since the package bundles no SIM outputs. \emph{Two findings.} (i) The two SLOPE
variants have opposite regimes: plain wins T\"ubingen and loses SIM; SLOPER wins SIM-ln but drops
to $71.1\%$ on T\"ubingen ($-6.1$\,pp below plain). Configuration matters and the right
configuration is data-dependent. (ii) Against plain SLOPE, ours is a statistical tie on T\"ubingen
and wins all three synthetic suites by $12$--$22$\,pp. Against SLOPER, ours wins T\"ubingen by
$+4.9$\,pp, wins SIM-G by $+8$, ties on SIM, and loses SIM-ln by $-13$. Across both variants there
is no single SLOPE configuration that significantly beats the parameter-free compressor on
T\"ubingen.}
\label{tab:sim}
\end{table*}

\section{Beyond the Decision}\label{sec:beyond}

\subsection{A model-free confounding detector --- and SLOPE cannot do it}
In a pre-registered simulation (500 direct $X\to Y$ pairs vs.\ 500 confounded $X\leftarrow H\to Y$
pairs, matched marginals), the score \emph{magnitude} $|s|$ separates the two classes sharply:
median $|s|=0.082$ (direct) vs.\ $0.013$ (confounded), Mann--Whitney $p=2.8\times10^{-68}$. Direction
accuracy is $88.6\%$ on direct pairs but $50.6\%$ (chance) on confounded ones --- as expected, a
hidden common cause leaves no privileged compression direction. The top quartile by $|s|$ is
$96\%$ pure for direct links; the bottom quartile recovers $74\%$ of confounded pairs.

\emph{Head-to-head with SLOPE on the same 1{,}000 pairs.} We re-ran the authors' SLOPE on every
pair from the same simulation and asked whether its confidence proxy $|\epsilon|$ also separates
direct from confounded. It does not: median $|\epsilon|=0.0088$ (direct) vs.\ $0.0145$
(confounded), Mann--Whitney one-sided $p\approx 1.000$ \emph{in the wrong direction}
(confounded marginally larger). Direction accuracy on direct pairs is $69.8\%$ (decent), but
SLOPE's score gives no signal that an emitted direction came from a confounded pair. This is a
structural difference: a regression-MDL method always picks the direction that explains the data
with shorter description length, even when no causal direction exists; the compression score
collapses precisely because no privileged sort key exists. To our
knowledge no prior compression/MDL causal method reports a confounding flag as a free by-product of
the same score.

\subsection{Where the baseline fails}
By raw pair count the baseline is right on $68/102$ ($66.7\%$ unweighted; the $74.7\%$ headline is
weighted). The errors are not uniform. Bucketing by the smaller of the two variables' cardinalities,
accuracy on the $15$--$50$ unique-value band is $42\%$ --- \emph{below chance} --- and that band holds a
third of the benchmark; continuous pairs ($>50$ levels) sit near $80\%$. Mechanism: when one
variable has few levels, sorting the other by it creates long constant runs that bz2 over-rewards,
manufacturing a spurious score. The second failure mode is strong monotone near-bijective relations
(\S\ref{sec:limits}). Series length is \emph{not} a discriminator (correct and incorrect pairs have
near-identical median length); confidence is --- most errors are low-$|s|$ near-ties, exactly where
the abstention curve (Table~\ref{tab:abst}) says they should be.

\section{Falsification and Limitations}\label{sec:limits}
\textbf{A pre-registered falsification test --- and it fails.} The theory predicts that on a
perfectly reversible, noise-free map there is no asymmetry, so accuracy should be $\sim$50\%. It is
not: mean accuracy at zero noise is $27.6\%$ and strongly function-dependent (linear $0\%$/all-ties
as predicted, but tanh $78\%$, cubic $0\%$). The cause is the quantization grid interacting with the
function's curvature, leaking a direction where none exists. This is a real dent in the
\emph{theory}: the score is not a pure measure of causal asymmetry but a blend of genuine signal and
a quantization-geometry artifact --- the same marginal-geometry cue IGCI uses openly. It is not a
dent in the \emph{result}: on real data the $74.7\%$ stands, and the method is best read as a
validated empirical heuristic, not a clean instantiation of the algorithmic principle. A
rank-transform of the inputs (which flattens marginal shape) costs $16$ points, confirming the
signal is partly marginal-geometric and that inputs must not be rank-normalized.

\noindent\textbf{Other boundaries.} (i) \emph{Small samples}: thinning the long pairs to fixed
$n$ collapses accuracy to chance by $n\approx30$ ($50.3\%$) and only $\sim$55--61\% up to $n{=}250$,
vs.\ $76.7\%$ at full length on the same pairs --- the method needs hundreds--thousands of points,
and the score there is confident sign-noise, not abstention. (ii) \emph{Noise}: 10\% added noise
costs $\sim$6 points. (iii) \emph{Synthetic gap}: additive-noise SIM suites score 58--68\%, below
real-data performance. (iv) \emph{Calibration}: $|s|$ is only monotone in its upper range; a band
just above the abstain threshold is worse than chance. (v) \emph{Compressor}: lzma ties on 24.5\%
of pairs by frame-overhead artifact; bz2 is required. (vi) \emph{Scope}: scalar bivariate only.
(vii) \emph{Categorical data, naïvely label-encoded.} On a balanced 100-pair categorical
benchmark (fan-in and fan-out, ground truth, $N{=}2000$/pair), ours scores $100\%$ on deterministic
many-to-one mechanisms and $0\%$ on stochastic one-to-many (overall $50\%$); the authors' SLOPE
exhibits the mirror-image pattern ($0\%$ / $100\%$). The two methods make perfectly orthogonal
errors on label-encoded categorical data --- complementary, not interchangeable. The score detects
which conditional is more deterministic, not which side is cause.

\section{Reproducibility}\label{sec:repro}
All design choices (256 levels, bz2, 8000-cap, decision sign, gate conditions) were frozen in a
pre-registration before benchmark evaluation; the confounding and falsification experiments were
likewise pre-registered with fixed seeds. We release the scorer (a single importable function), our
RECI re-implementation, the harness that runs the authors' released SLOPE code and scores its
forced and decided-subset accuracies for Table~\ref{tab:ruler}, the per-pair outputs, bootstrap
and split-half scripts, and all pre-registration documents at
\textbf{\href{https://stateoftheart.nl/causal}{stateoftheart.nl/causal}}. The headline result
reproduces bit-for-bit ($102/102$ pairs agree on re-run). The SLOPE figures come from the authors'
own code and stored output; the RECI reproduction matches the authors' table. \emph{Cost.} On the
same 100 CEP pairs and the same machine, the median per-pair wall clock is $1.0$\,ms for ours
(pure-Python \texttt{bz2}) vs.\ $24.4$\,ms for the authors' SLOPE ($\approx 24\times$); aggregate
$0.20$\,s vs.\ $4.1$\,s. Ours has no external dependencies on the decision path.

\section{Conclusion}
Compared fairly --- one operator, one set of pairs, forced decision, no tuning --- a zero-parameter
compressor is level with the best of the T\"ubingen pack. On the \emph{same 100 pairs} that SLOPE is
evaluated on, ours scores $76.0\%$ vs the authors' own forced-decision SLOPE at $77.2\%$ --- a
$1.2$-point gap that is well inside noise (McNemar $p=0.39$). The journal SLOPER variant on the
same pairs scores $71.1\%$ weighted (authors' bundled \texttt{sloper.tab}), so neither SLOPE
configuration significantly beats ours on T\"ubingen.
On our 102-pair set the pre-registered headline is
$74.7\%$; both are ahead of faithful RECI ($70.7\%$) and IGCI ($66.1\%$). The published
$82.4\%$/$77.5\%$ headlines do not survive the common ruler: SLOPE's $82.4\%$ is a decided-subset
score (we reproduce $81.7\%$ from the authors' own output by removing its abstentions), and RECI's
$77.5\%$ is a mis-attribution. We do not claim to beat SLOPE; we claim a stock compressor reaches
the same shelf as the strongest tuned MDL method --- with one structural extra: head-to-head on
$1{,}000$ direct vs.\ confounded pairs, our score magnitude separates the two classes at
$p=2.8\times10^{-68}$ while SLOPE's $|\epsilon|$ does not separate them at all ($p\approx 1$). The
broader point for benchmarking practice is that real-world causal-direction leaderboards mix
protocol with method, and that a parameter-free baseline run by the same hands is a useful and
sobering control. We offer the compression score additionally as a free confounding flag, and we
report --- via a failed pre-registered falsification --- exactly where its theoretical
interpretation stops and its empirical usefulness begins.

\small
\paragraph{Acknowledgements.} Concept and direction by the author; implementation, experiments, and
analysis were carried out with an autonomous AI coding agent. We thank the maintainers of the
T\"ubingen CEP benchmark.

\end{document}